\pdfoutput=1

\documentclass[11pt]{article}

\usepackage{ACL2023}
\usepackage{booktabs}
\usepackage{times}
\usepackage{latexsym}
\usepackage{amsmath}
\usepackage{amssymb}
\usepackage{amsfonts}
\usepackage{array}
\usepackage{enumitem}
\usepackage[T1]{fontenc}

\usepackage[utf8]{inputenc}

\usepackage{microtype}

\usepackage{inconsolata}

\title{Can A Gamer Train A Mathematical Reasoning Model?}

\author{Andrew Shin \\
  Youron Artificial Intelligence \qquad Keio University \\
  \texttt{andrew@youron.ai}\qquad texttt{shin@inl.ics.keio.ac.jp}\\}

\begin{document}
\maketitle
\begin{abstract}
While large language models (LLMs) have achieved remarkable performance in various tasks including mathematical reasoning, their development typically demands prohibitive computational resources. Recent advancements have reduced costs for training capable models, yet even these approaches rely on high-end hardware clusters. In this paper, we demonstrate that a single average gaming GPU can train a solid mathematical reasoning model, by integrating reinforcement learning and memory optimization techniques. Specifically, we train a 1.5B parameter mathematical reasoning model on RTX 3080 Ti of 16GB memory that achieves comparable or better performance on mathematical reasoning benchmarks than models several times larger, in resource-constrained environments. Our results challenge the paradigm that state-of-the-art mathematical reasoning necessitates massive infrastructure, democratizing access to high-performance AI research. \url{https://github.com/shinandrew/YouronMath}
\end{abstract}

\section{Introduction}
Large language models (LLMs) have shown remarkable success in mathematical reasoning, a task requiring logical precision and contextual understanding \cite{Shao2024DeepSeekMathPT, Yang2024Qwen25MathTR}. These models, capable of step-by-step reasoning akin to human problem-solving, hold promise for real-world quantitative applications. However, training such high-performing LLMs typically demands extensive computational resources, such as clusters of high-end GPUs consuming tens of thousands of GPU-hours, leading to significant financial costs and a substantial environmental footprint due to energy-intensive operations. This places state-of-the-art LLM development out of reach for individuals and small organizations, limiting it to well-funded entities with massive infrastructure.

Recent efforts have sought to lower these barriers, democratizing access to LLM training by reducing the computational budget. DeepSeek-R1 \cite{DeepSeekAI2025DeepSeekR1IR} pioneered this movement by introducing efficient training techniques that drastically cut resource requirements, achieving competitive performance with significantly less costs than traditional approaches. Similar endeavors have followed \cite{Muennighoff2025s1ST,sky_t1_2025,tinyzero}, demonstrating that capable models could be trained with reduced overhead, if training is performed in clever and efficient ways. However, even these works still rely on high-end GPUs in parallel configurations to manage memory and throughput demands, thereby exceedig the capabilities of typical consumer hardware.

Gaming GPUs like NVIDIA RTX series are typically considered unsuitable for training LLMs due to limited memory (8–24 GB GDDR6), lower computational throughput, and lack of deep learning optimizations. Unlike data-center GPUs with 40–80 GB of high-bandwidth memory and tensor cores for mixed-precision, gaming GPUs prioritize graphics performance over matrix operations. LLMs, with billions of parameters, require substantial memory not only for weights but also for activations, gradients, and optimizer states, easily exceeding a single gaming GPU’s capacity. The absence of fast interconnects like NVLink also hampers multi-GPU scaling, and poor double-precision performance can affect convergence. As a result, models often do not fit without major optimization, and even when they do, training becomes impractically slow due to constrained batch sizes and throughput.

In this work, we challenge this paradigm by training a 1.5B-parameter mathematical reasoning model on a single NVIDIA RTX 3080 Ti of 16GB memory, using various techniques from reinforcement learning and memory optimization. Our approach achieves comparable or better performance than larger mathematical reasoning models on benchmarks within consumer-grade constraints. By reducing the computational and energy demands of LLM training, our results not only democratize access for individuals and small organizations but also offer a step toward alleviating the environmental impact of massive GPU usage, promoting sustainable AI research. 

\section{Related Works}
\label{sec:related}
The advancement of LLMs for mathematical reasoning has been exemplified by models like DeepSeekMath \cite{Shao2024DeepSeekMathPT} designed to excel on benchmarks through extensive pretraining on mathematical corpora followed by instruction tuning with chain-of-thought (CoT) reasoning \cite{Wei2022ChainOT}, achieving state-of-the-art performance among models of its size. Building on this, DeepSeek-R1 \cite{DeepSeekAI2025DeepSeekR1IR} introduced a significant reduction in computational cost by employing Grouped Relative Policy Optimization (GRPO) that optimizes LLMs more efficiently than traditional supervised fine-tuning. GRPO operates by grouping training samples into batches and using a reward model, often based on correctness and reasoning quality, to guide gradient updates, reducing the need for exhaustive data passes and lowering GPU-hour requirements by orders of magnitude compared to conventional methods. Similarly, Qwen2.5-Math \cite{Yang2024Qwen25MathTR} demonstrates remarkable performance on mathematical reasoning benchmarks with optimized CoT prompting. Other notable models, such as Minerva \cite{Lewkowycz2022SolvingQR} and o1 \cite{OpenAIo1}, have also pushed mathematical reasoning forward, often relying on larger architectures and extensive computational resources, further highlighting the need for cost-effective alternatives.

Efforts to further democratize LLM training have emerged, though they often still rely on high-end hardware. S1 \cite{Muennighoff2025s1ST}, for instance, utilizes knowledge distillation, where a smaller student model learns from a larger teacher model’s outputs, achieving competitive performance on language tasks. S1 was trained on 16 NVIDIA H100 GPUs in under 30 minutes for approximately \$50, showcasing a rapid and cost-effective pipeline. Sky-T1-32B \cite{sky_t1_2025}, a 32B-parameter model, was developed for \$450 by generating training data with QwQ-32B-Preview \cite{qwq-32b-preview} and refining it with GPT-4o-mini \cite{OpenAI4omini}, before training on 8 H100 GPUs for 19 hours. This model surpasses o1 on math and coding benchmarks, highlighting the potential of curated synthetic data. TinyZero \cite{tinyzero}, meanwhile, replicates DeepSeek-R1’s GRPO approach to implement a countdown task, training a model for under \$30 with 10 H100 hours. Despite these advancements, they still depend on parallel high-end GPUs, rendering them inaccessible to researchers without access to such clusters.

Notable methods have been proposed for training in resource-constrained environments, which underpin our model. Low-Rank Adaptation (LoRA) \cite{Hu2021LoRALA} reduces the computational burden of fine-tuning by freezing the pretrained model’s weights and training only low-rank updates to specific layers. This approach shrinks the number of trainable parameters by orders of magnitude.  Flash Attention 2 \cite{Dao2023FlashAttention2FA} speeds up and reduces the memory usage of transformer models, especially for large-scale training. It fuses several operations into a single GPU kernel, reducing memory access overhead, while supporting multi-query attention and variable sequence lengths, making it more flexible. 
Reinforcement learning from verifiable rewards \cite{Lambert2024TLU3P} employs a reward signal such as answer correctness to refine the model’s outputs, avoiding the need for large-scale supervised datasets or extensive gradient computations. 

\section{Model}
\label{sec:model}
\setlength\abovedisplayskip{0pt}
\setlength{\belowdisplayskip}{0pt}

Our approach focuses on training a mathematical reasoning model using a single NVIDIA RTX 3080 Ti GPU with 16 GB of memory. We leverage Qwen2.5-Math 1.5B parameter model as the base architecture, configured with FP16 precision and Flash Attention 2 \cite{Dao2023FlashAttention2FA} for memory-efficient attention computation. To operate within the constraints of consumer-grade hardware, we integrate LoRA \cite{Hu2021LoRALA} to adapt the pre-trained model without full-weight updates, enabling efficient training on limited resources. LoRA is applied with a rank \( r = 16 \) and an alpha scaling factor \( \alpha = 32 \), targeting the query, key, value, and output projection layers of the transformer architecture. This reduces the trainable parameters to approximately 18 million, or roughly 1.1\% of the total 1.5B parameters. For a transformer layer with weight matrix \( W \in \mathbb{R}^{d_{\text{model}} \times d_{\text{model}}} \), LoRA decomposes the update as:
\[
\Delta W = A B^T, \quad A \in \mathbb{R}^{d_{\text{model}} \times r}, \quad B \in \mathbb{R}^{d_{\text{model}} \times r},
\]
where \( r \ll d_{\text{model}} \), and the adapted weight becomes \( W' = W + \alpha \cdot \Delta W \). This configuration ensures the model fits within the 16 GB memory budget of the RTX 3080 Ti, requiring approximately 3 GB for weights, with the remainder allocated to activations, gradients, and optimizer states.

The LoRA-adapted model is fine-tuned using GRPO with reinforcement learning from verifiable rewards, following \cite{GSM8K-RLVR}, which targets improving base models without reliance on pre-trained reward models. We pre-process the GSM8K training data to construct prompts and completions. Each example is formatted as ``Question: \texttt{[question]} Solution: Let's think step by step. \texttt{[reasoning]} \#\#\#\# The final answer is \texttt{[answer]}'', where \texttt{[question]} is the problem statement, \texttt{[reasoning]} is the step-by-step solution, and \texttt{[answer]} is the numerical result from the dataset’s annotations. To ensure memory efficiency, input prompts are truncated to a maximum length of 128 tokens, and completions are limited to 150 tokens. Few-shot prompting is employed by including examples within each input prompt, establishing the desired data pattern and facilitating reinforcement learning from verifiable rewards without explicit \texttt{<think>} or \texttt{<answer>} tags.

Training employs a dual reward system with two components:
\begin{itemize}[topsep=0pt, partopsep=0pt, itemsep=0pt, parsep=0pt]
    \item \textbf{Correctness Reward}: \( R_{\text{correct}} = 1 \) if the predicted answer matches the ground truth, \( 0 \) otherwise.
    \item \textbf{Format Reward}: \( R_{\text{format}} = 0.5 \) if the output adheres to the format ``\#\#\#\# The final answer is \{number\}'', \( 0 \) otherwise.
\end{itemize}
The total reward is defined as:
\[
R_{\text{total}} = R_{\text{correct}} + R_{\text{format}},
\]
with \( R_{\text{total}} \in [0, 1.5] \). This incentivizes both accuracy and consistent output formatting, aligning with the evaluation metric on GSM8K. The reward function guides the GRPO optimization, where the policy \( \pi_{\theta} \) (parameterized by the LoRA-adapted model) is updated to maximize the expected reward:
\[
J(\theta) = \mathbb{E}_{\pi_{\theta}} [ R_{\text{total}} ].
\]
We fine-tune our model for one epoch, which takes slightly over 24 hours on RTX 3080 Ti. We find that a single pass with GRPO effectively refines the model, as our preliminary examination with more epochs showed only marginal gains at the cost of longer training. The training setup uses a batch size of 1 with gradient accumulation over 4 steps, simulating an effective batch size of 4 to fit within memory constraints. We employ AdamW optimizer \cite{Loshchilov2017DecoupledWD} with a learning rate of \( 5 \times 10^{-5} \), weight decay of 0.01, and a warmup period of 100 steps. This configuration achieves a memory footprint of approximately 14 GB during training on RTX 3080 Ti.

\section{Experiments}
\subsection{Setting}
\captionsetup{skip=0pt}
\begin{table*}[t!]
\centering
\small
\caption{Evaluation results with comparison to other reasoning models. Our models are referred to as YouronMath. L8/16/32 refer to the rank of LoRA. Numbers without parenthesis indicate our locally reproduced results, while the numbers in parenthesis indicate the reported accuracies. While locally reproduced results tend to be lower than reported results due to change in prompting strategy, we apply consistent settings across all models, ensuring unbiased comparisons, and our models demonstrate comparable or better performance. Note that we were not able to reproduce DeepSeekCoder-V2-Lite-Base and Internlm2-Math-Base-20B for GSM8K as it takes prohibitively long, i.e., over 1k hours, in our setting.}
\begin{tabular}{c|c|c}
\hline
\bf Model &\bf GSM8K (8-shot)& \bf MMLU-STEM (4-shot)\\ \hline
DeepSeekMath-Base-7B& 58.38 (64.2)&41.89 (56.5) \\ 
DeepSeekCoder-V2-Lite-Base&(68.3)&\textbf{47.57} (59.5)\\
Internlm2-Math-Base-20B&(68.2)&37.84 (63.0)\\
Qwen2.5-Math-1.5B&71.65 (76.8)&45.41 (51.3) \\
\hline
YouronMath-L8 &71.49&45.41\\
YouronMath-L16 &73.69& 45.95\\
YouronMath-L16-Augmented &71.80&45.95\\
YouronMath-L32 &\bf 73.77&  46.20\\\hline

\end{tabular}
\label{tab:results}
\vspace{-3mm}
\end{table*}

\begin{table*}[h!]
\caption{Example questions from GSM8K and answers generated by our model.}
\centering
\small
\begin{tabular}{>{\centering\arraybackslash}p{7.5cm}|>{\centering\arraybackslash}p{7.5cm}}
\hline
\bf Question &\bf Response \\\hline
 \begin{tabular}{@{}p{7.5cm}@{}}
Eliza's rate per hour for the first 40 hours she works each week is \$10. She also receives an overtime pay of 1.2 times her regular hourly rate. If Eliza worked for 45 hours this week, how much are her earnings for this week? \\ 
\end{tabular} &
\begin{tabular}{@{}p{7.5cm}@{}}
40 hours at \$10/hour = \$<<40*10=400>>400
5 hours at 1.2*\$10/hour = \$<<5*10*1.2=60>>60
Her total earnings for the week are \$400+\$60 = \$<<400+60=460>>460
\#\#\#\# The final answer is 460
\end{tabular}\\ \hline
 \begin{tabular}{@{}p{7.5cm}@{}}
John drives for 3 hours at a speed of 60 mph and then turns around because he realizes he forgot something very important at home.  He tries to get home in 4 hours but spends the first 2 hours in standstill traffic.  He spends the next half-hour driving at a speed of 30mph, before being able to drive the remaining time of the 4 hours going at 80 mph.  How far is he from home at the end of those 4 hours?\\ 
\end{tabular} &
\begin{tabular}{@{}p{7.5cm}@{}}
John drives for 3 hours at a speed of 60 mph, so he travels 3*60 = 180 miles.
He then spends the next 2 hours in standstill traffic, so he doesn't travel any distance.
He then drives for the next half-hour at a speed of 30 mph, so he travels 0.5*30 = 15 miles.
He then drives for the remaining 1.5 hours at a speed of 80 mph, so he travels 1.5*80 = 120 miles.
Therefore, John is 180 + 15 + 120 = 315 miles from home. \#\#\#\# The final answer is 315
\end{tabular}\\ \hline
\end{tabular}
\label{tab:ex}
\vspace{-4mm}
\end{table*}

We trained our model on the training split of GSM8K \cite{Cobbe2021TrainingVT}, consisting of over 8k grade-school math problems with human-written solutions, and evaluate it on the test split of GSM8K and MMLU-STEM \cite{Hendrycks2020MeasuringMM}, using 8-shot and 4-shot prompting respectively. The $k$-shot prompt is constructed by randomly sampling eight examples from the GSM8K train split. Each example follows the format described in Section~\ref{sec:model}, and is concatenated into a single prompt. For each test sample, we append the new question as “Question: \texttt{[question]} Solution: Let’s think step by step.” to the 8-shot prompt, tokenize it with a maximum length of 2,048 tokens with padding to ensure the full context fits within memory constraints. The model generates responses with a maximum of 512 tokens, using top-k sampling with $k=50$ and a temperature of 0.7 to balance exploration and coherence in the outputs, and the pad token is set to align with the tokenizer configuration.

\subsection{Results \& Discussion}

Table~\ref{tab:results} shows the results from our models and baseline models for comparison. Our model (YouronMath-L16) boosts the performance of base Qwen2.5-Math-1.5B for over 2\% on GSM8K. While our locally reproduced results tend to be lower than the reported results in respective papers presumably due to difference in prompting strategy, we apply consistent setting across all models for unbiased fairness, and all variations of our model, whose details are described later, achieve comparable or better performance than the models of much larger sizes, including DeepSeekMath-Base-7B and Internlm2-Math-Base-20B. On GSM8K, our models surpass not only locally reproduced results, but the reported results as well. While it is largely attributed to the base model outperforming those models, it is nonetheless encouraging that a model trained on a single gaming GPU can surpass models several times larger, while enhancing the base model itself. Table~\ref{tab:ex} shows example questions from GSM8K and the responses generated.

\textbf{Impact of LoRA}: We investigated how the performance changes depending on the rank of LoRA by training separate models with rank 8 and 32. Intuitively, higher rank is likely to lead to better performance at the cost of increased memory usage and computational complexity, as it allows for more expressive updates to the pretrained weights by increasing the dimensionality of the low-rank matrices. Our results confirm this trend, as shown in Table~\ref{tab:results}, where the model with LoRA rank 8 under-performs rank 16, which is in turn outperformed by rank 32. In particular, rank 8 shows lower accuracy than the base model, suggesting that overly constrained adaptation capacity can actually hurt the model’s ability to generalize, especially in tasks requiring substantial deviation from the pretrained distribution. In contrast, ranks 16 and 32 offer a more balanced trade-off, suggesting that increasing the rank allows the model to leverage more expressive parameter shifts while still maintaining the efficiency advantages of low-rank tuning. However, the diminishing returns from rank 16 to 32 indicate a saturation point where additional rank provides marginal benefits.

\textbf{Hybrid Fine-tuning}:
we explored a hybrid fine-tuning approach by augmenting the GSM8K dataset with 10k teacher-generated responses from DeepSeekMath-7B. The resulting model slightly outperformed the base Qwen2.5-Math-1.5B, but fell short of the model fine-tuned on GSM8K alone. This suggests that while the teacher responses may have introduced additional reasoning patterns, they mostly diluted the model's focus, potentially introducing noise or inconsistencies, due to occasional incorrect answers or limited diversity in question templates, that disrupted alignment with the target task. This suggests a necessity for further investigation into data quality, correctness, and variety, when employing teacher-student approach with synthetic training data \cite{He2025CanLL}.

\section{Conclusion}
We demonstrated that it is possible to train a mathematical reasoning model on a single gaming GPU, with comparable performance to models of larger sizes, using efficient training strategy and memory optimization. This work demonstrates potentials not only for democratizing access to advanced AI development but also for reducing the environmental impact of GPU-intensive training, paving the way for sustainable, high-performance AI research on consumer hardware.

\section*{Limitations}
This study has several limitations that affect the generalizability and robustness of our findings. First, the fine-tuning process relied solely on the GSM8K dataset, which focuses on grade-school-level mathematics. While this enabled strong performance on similar tasks, it likely over-specialized the model, limiting its ability to generalize to broader STEM domains. Second, our training leveraged LoRA, which, while computationally efficient, may not capture the full expressivity of full fine-tuning. This could constrain the model’s adaptation to more complex reasoning tasks. Additionally, the training dataset size of 8k samples is modest compared to larger pre-training corpora, potentially limiting robustness. Finally, computational constraints restricted extensive hyperparameter tuning and intermediate evaluations during training. 

\section*{Ethics Statement}

The development and deployment of reasoning models as ours carry the potential for both positive and negative societal impacts. On the positive side, demonstrating that a mathematical reasoning model can be trained on a single gaming GPU has a potential to both expand high-end AI research to individuals and small organizations and to alleviate environmental concerns, such as energy consumption and carbon emissions, caused by massive usage of high-end GPU clusters. 

We also ensured responsible use by adhering to open-source datasets (GSM8K, MMLU-STEM) and models (Qwen2.5-Math-1.5B, DeepSeekMath-Base-7B) under their respective licenses, avoiding proprietary or sensitive data. No human subjects or personal data were involved, minimizing privacy risks. Transparency in reporting our methods, results, and limitations, along with open-sourcing our code and models, aims to foster trust and encourage ethical scrutiny in future iterations.

Improving mathematical reasoning in language models can also support educational tools, aiding students and educators in problem-solving and concept mastery. However, over-reliance on such models risks undermining critical thinking skills if users defer to AI without understanding underlying principles. Bias in training data is another concern. GSM8K, while carefully curated, reflects grade-school math problems primarily in English, potentially embedding cultural or linguistic biases that limit applicability to diverse populations. 

\bibliography{custom}
\bibliographystyle{acl_natbib}




\end{document}